\documentclass{article}

\PassOptionsToPackage{numbers, compress, sort}{natbib}

\usepackage{graphicx}
\usepackage{tabularx}
\usepackage{array} 
\usepackage{adjustbox}
\usepackage[most]{tcolorbox}
\usepackage{setspace}
\usepackage{stackengine}
\usepackage{caption}
\usepackage{etoolbox}
\usepackage{textgreek}
\usepackage[htt]{hyphenat}

\newtoggle{anonymous}
\togglefalse{anonymous}
\iftoggle{anonymous}{\usepackage{iclr2027_conference}}{\usepackage{iclr2027_conference} \iclrpreprintcopy}

\usepackage[utf8]{inputenc} % allow utf-8 input
\usepackage[T1]{fontenc}    % use 8-bit T1 fonts
\usepackage{hyperref}       % hyperlinks
\usepackage{url}            % simple URL typesetting
\usepackage{booktabs}       % professional-quality tables
\usepackage{amsfonts}       % blackboard math symbols
\usepackage{nicefrac}       % compact symbols for 1/2, etc.
\usepackage{microtype}      % microtypography
\usepackage{xcolor}         % colors
\usepackage{amsmath}
\usepackage{amssymb}
\usepackage{pifont}
\usepackage{seqsplit}
\usepackage{textcomp}

\DeclareMathOperator*{\E}{\mathbb{E}}
\DeclareMathOperator*{\argmin}{\mathrm{argmin}}
\DeclareMathOperator*{\argmax}{\mathrm{argmax}}
\newcolumntype{Y}{>{\centering\arraybackslash}X}
\newcommand{\R}{\mathbb{R}}
\newcommand{\methodfull}{Smooth Tchebycheff Optimization of Multi-Objective Preferences}
\newcommand{\method}{STOMP}

\title{Pareto-Optimal Offline Reinforcement Learning via Smooth Tchebycheff Scalarization}

\renewcommand{\thefootnote}{\fnsymbol{footnote}}
\author{Aadyot Bhatnagar\footnotemark[1] \textsuperscript{,}\footnotemark[3]
\And
Peter M{\o}rch Groth\footnotemark[1]
\And
Sebastian Ibarraran\footnotemark[1] \textsuperscript{,}\footnotemark[2]
\And
Ali Madani\footnotemark[1] \textsuperscript{,}\footnotemark[3]
}

\begin{document}

\maketitle
\iftoggle{anonymous}{}{
\footnotetext[1]{Profluent Bio, Inc.}
\footnotetext[2]{Stanford University; work done as an intern at Profluent Bio, Inc.}
\footnotetext[3]{To whom correspondence should be addressed: \texttt{\{abhatnagar,ali\}@profluent.bio}}
}
\renewcommand{\thefootnote}{\arabic{footnote}}
\setcounter{footnote}{0} 
\begin{abstract}
Large language models can be aligned with human preferences through offline reinforcement learning (RL) on small labeled datasets.
While single-objective alignment is well-studied, many real-world applications demand the simultaneous optimization of multiple conflicting rewards, e.g.\ activity and specificity for proteins, or helpfulness and harmlessness for chatbots.
Prior work has largely relied on linear reward scalarization, which provably fails to recover non-convex regions of the Pareto front.
In this paper, instead of scalarizing the rewards directly, we frame multi-objective RL itself as an optimization problem to be scalarized via smooth Tchebycheff scalarization, a recent technique that overcomes the drawbacks of linear scalarization. 
We use this formulation to derive Smooth Tchebycheff Optimization of Multi-Objective Preferences (STOMP), a novel offline RL algorithm that extends direct preference optimization to the multi-objective setting in a principled way by standardizing the individual rewards based on their observed distributions.
We empirically validate STOMP using multiple base models on 3 protein engineering and 3 chatbot alignment datasets.
Compared to state-of-the-art baselines, STOMP achieves or ties for the highest hypervolumes on 16/18 protein tasks and 5/6 natural language tasks.
We thus demonstrate that STOMP is a powerful, robust multi-objective alignment algorithm that can meaningfully improve post-training in multiple domains.
\end{abstract}

\section{Introduction}
Large language models have been trained on massive corpora of text, code, images, and proteins \citep{vaswani2017attention, madani2023progen, meier2021esm1v, brown2020gpt3, devlin2019bert, radford2021clip, dosovitskiy2021vit}. These models are powerful density estimators for the data distributions they were trained on, but they often require supervised post-training via algorithms like reinforcement learning from human feedback (RLHF) to {\em align} their likelihoods with the preferences of their users \citep{ouyang2022rlhf, ziegler2020rlhf, christiano2017rlhf}. RL-based alignment is an incredibly general paradigm that has been used to train chatbots \citep{ouyang2022rlhf, bai2022anthropichh, rafailov2023dpo}, improve text-to-image generators \citep{liang2024rlhfimage, lee2024parrot}, and enable protein language models to generate proteins with higher fitness \citep{bhatnagar2025progen3, ibarraran2026efficient, wen2025paretooptimal, widatalla2024proteindpo}.

While single-objective RL is well-studied, many real-world applications demand the simultaneous optimization of multiple conflicting objectives. For example, chatbots should be helpful to their users, but their responses should also be safe \citep{bai2022anthropichh}. Text-to-image models should generate high-quality images that are aligned with their prompts \citep{lee2024parrot}. Protein engineers may seek a protein with high activity and specificity \citep{wang2025activelearningde}, or a highly active protein that maintains adequate stability and expression \citep{armer2025protein-eng-tournament}. Since no single solution can in general optimize all these rewards simultaneously, our goal is instead the Pareto front, i.e.\ the set of all non-dominated solutions.

Scalarizing the reward vector is a common approach to multi-objective RL (MORL), with linear scalarization (i.e.\ taking a weighted average) being the simplest strategy \citep{li2021morl, zhou2024modpo, wen2025paretooptimal, chennakesavalu2025aligning, yang2019morl}. Unfortunately, linear reward scalarization is provably incapable of recovering solutions from non-convex regions of the Pareto front, and solutions from these regions often represent the exact compromises between conflicting objectives that we are the most interested in discovering \citep{das1997closer, boyd2004convex, ehrgott2005multicriteria}.

\citet{lin2024smooth} introduce smooth Tchebycheff scalarization (STS) to address the shortcomings of linear scalarization. STS enjoys efficient gradient-based optimization and is able to find all solutions on the Pareto front. However, STS can be challenging to use off-the-shelf for reward scalarization because it introduces many free hyperparameters \citep{qiu2024traversingparetooptimalpolicies}.
In this work, rather than using STS to scalarize the rewards directly, we observe that MORL is itself a multi-objective optimization problem to be scalarized. We thus apply STS to the MORL problem and use this framing to derive a reward scalarization that dynamically standardizes the rewards based on their observed distributions in an offline training set.

Building on these insights, we introduce \methodfull{} (\method{}), an extension of direct preference optimization \citep{rafailov2023dpo} that aligns language model likelihoods using an offline dataset that associates each sequence to vector-valued rewards. We empirically evaluate \method{} by aligning three protein language models on three experimental laboratory datasets of protein fitness, and two instruction-tuned language models on three chatbot alignment datasets. Compared to state-of-the-art baselines that use either linear scalarization or a more naive implementation of STS, we find that \method{} achieves (or ties for) the highest hypervolumes in 16/18 protein evaluation settings and 5/6 natural language evaluation settings.

\section{Preliminaries and Related Work}
\subsection{Reinforcement Learning}
Given context $x \in \mathcal{X}$, a language model or policy $\pi$ predicts the conditional probability $\pi(y | x)$ of sequences $y \in \mathcal{Y}$. Most reinforcement learning (RL) formulations for language models optimize the expected reward of sequences generated from $\pi$ while simultaneously requiring that $\pi$ does not deviate too much from a reference $\pi_0$, i.e.\ $\mathbb{D}_{\mathrm{KL}}(\pi \mid\mid \pi_0) \le \eta$. $\pi_0$ is typically the pre-trained or fine-tuned language model that we are further training with RL \citep{schulman2015trpo, schulman2017ppo, christiano2017rlhf, ziegler2020rlhf}. Concretely, given a distribution over contexts $\mathcal{P}$ and a reward function $r: \mathcal{X} \times \mathcal{Y} \to \R$, we seek to solve
\begin{align*}
& \argmax_{\pi} \E_{\substack{x \sim \mathcal{P} \\ y \sim \pi(\cdot \mid x)}} [r(x, y)] \text{\quad s.t. \quad} \mathbb{D}_{\mathrm{KL}}(\pi \mid\mid \pi_0) \le \eta \\
=& \argmax_{\pi} \E_{\substack{x \sim \mathcal{P} \\ y \sim \pi(\cdot \mid x)}} [r(x, y)] - \beta \mathbb{D}_{\mathrm{KL}}(\pi \mid\mid \pi_0) \\
=& \argmin_{\pi} \E_{\substack{x \sim \mathcal{P} \\ y \sim \pi(\cdot \mid x)}} \left[\beta \log \pi(y | x) - \beta \log \pi_0(y | x) - r(x, y) \right] \\
=& \argmin_{\pi} \mathbb{D}_{\mathrm{KL}}(\pi \mid\mid \pi^\star)
\end{align*}
\citet[Lemma 5]{faury2020distributionallyrobust} prove that the optimal Lagrange multiplier is $\beta = \sqrt{\mathbb{V}_{\pi_0}[r(x, y)] / 2 \eta}$. Meanwhile, $\pi^\star$ is the optimal policy and $Z(x)$ is its partition function:
\begin{align*}
\pi^\star(y | x) = \frac{1}{Z(x)} \pi_0(y | x) \exp\left(r(x, y) / \beta \right) \quad \text{and} \quad Z(x) = \sum_{y \in \mathcal{Y}} \pi_0(y | x) \exp\left(r(x, y) / \beta \right)
\end{align*}
Maximum-entropy reinforcement learning (MaxEntRL) uses a uniform reference $\pi_0(y | x) \propto 1$ \citep{toussaint2009maxentrl, ziebart2008maxentinvrl}. MaxEntRL favors more stochastic policies with more diverse generations, and it has certain robustness guarantees \citep{eysenbach2022maxentrl-robust}. 

\subsection{Offline Reinforcement Learning via Preference Optimization}
RL is challenging because it requires us to optimize the expected reward as sampled from the policy $\pi$ we are training. Direct preference optimization (DPO) \citep{rafailov2023dpo} solves $r(x, y) = \beta \log (\pi^{\star}(y | x) / \pi_0(y | x)) + \beta \log Z(x)$ and assumes the Bradley-Terry model \citep{bradleyterry1952} $p(y_1 \succ y_2 | x) = \sigma(r(x, y_1) - r(x, y_2))$ to recast this challenging RL problem as the more tractable maximum likelihood estimation problem of minimizing the following loss on an offline dataset:
\begin{align}
    \mathcal{L}_{\mathrm{DPO}}(\pi; \beta, \delta) = \!\!\!\!\!\!\!\!\! \sum_{\substack{x, y_w, y_l \\ r(x, y_w) > r(x, y_l) + \delta}} \!\!\!\!\!\!\!\!\! -\log \sigma\left( \beta \log \frac{\pi(y_w | x)}{\pi_0(y_w | x)} - \beta \log \frac{\pi(y_l | x)}{\pi_0(y_l | x)} \right) \label{eq:dpo}
\end{align}
Here, $y_w$ is a ``winner'' and $y_l$ is a ``loser'' according to the reward $r$, while $\delta$ ensures that we only compare samples with a sufficiently large reward difference. SimPO \citep{meng2024simpo}, CPL \citep{hejna2024cpl}, and $\alpha$-DPO \citep{wu2025alphadpo} use a uniform reference $\pi_0(y | x) \propto 1$ and can be considered instantiations of MaxEntRL in the DPO framework.

Recent works enhance DPO by explicitly using the reward difference $r(x, y_w) - r(x, y_l)$ as a margin into similar paired losses \citep{chennakesavalu2025aligning, gao2024rebel, mao2024vcb, fisch2025robust, amini2024offsetdpo}. These methods leverage the difference
\begin{align*}
    \beta \log \frac{\pi(y_w | x)}{\pi(y_l | x)} \!-\! \beta \log \frac{\pi^{\star}(y_w | x)}{\pi^{\star}(y_l | x)} &= \Big(\!\underbrace{\beta \log \frac{\pi(y_w | x)}{\pi_0(y_w | x)} \!- r(x, y_w)}_{S^{\pi}(x, y_w; \beta)} \!\Big) \!-\! \Big(\! \underbrace{\beta \log \frac{\pi(y_l | x)}{\pi_0(y_l | x)} \! - r(x, y_l)}_{S^{\pi}(x, y_l; \beta)} \!\Big)
\end{align*}
to match the learned $\pi$ to the optimal $\pi^\star$ and can be theoretically grounded in various ways. For example, squared loss methods \citep{gao2024rebel, fisch2025robust, mao2024vcb} minimize $(S^{\pi}(x, y_w; \beta) - S^{\pi}(x, y_l; \beta) + \delta)^2$, while OffsetDPO \citep{amini2024offsetdpo} minimizes
\begin{align}
    \mathcal{L}_{\mathrm{ODPO}}(\pi; \beta, \delta) = \!\!\!\!\!\!\!\!\! \sum_{\substack{x, y_w, y_l \\ r(x, y_w) > r(x, y_l) + \delta}} \!\!\!\!\!\!\!\!\! -\log \sigma\left( S^{\pi}(x, y_w; \beta) - S^{\pi}(x, y_l; \beta) + \delta \right) \label{eq:odpo}
\end{align}

\subsection{Multi-Objective Optimization}
Multi-objective optimization minimizes $k$ distinct losses $\boldsymbol{L} = L_1, \ldots, L_k$ simultaneously. Often, these losses conflict with each other and cannot be simultaneously optimized by a single solution. We instead say that $\boldsymbol{L}(z') \prec \boldsymbol{L}(z)$ (i.e.\ $z'$ {\em Pareto-dominates} $z$) if $L_i(z') \le L_i(z)$ for all $i$ and $L_i(z') < L_i(z)$ for at least one $i$. The {\em Pareto front} of optimal solutions is the set of non-dominated solutions, i.e.\ $\{ z \in \mathcal{Z} : \boldsymbol{L}(z') \not\prec \boldsymbol{L}(z) \ \forall \ z' \in \mathcal{Z}\}$ \citep{miettinen1999}.

Scalarization is a common approach to multi-objective optimization, with linear scalarization being the simplest strategy, i.e.\ $\min_z \sum_{i=1}^{k} \lambda_i L_i(z)$ for some preference vector $\lambda \in \Delta^{k-1}$ on the simplex \citep{miettinen1999, geoffrion1967linearscalarization}. Without loss of generality, we assume $\lambda_i > 0$ for all $i$. However, linear scalarization will fail to find any solutions from non-convex regions of the Pareto front \citep{boyd2004convex, das1997closer, ehrgott2005multicriteria}.

Tchebycheff scalarization (TS) lets $L_i^\star = \min_z L_i(z)$ and solves $\min_z \max_i \lambda_i (L_i(z) - L_i^\star)$. Under mild assumptions, it can find all solutions on the Pareto front by varying $\lambda$ \citep{bowman1976Tchebycheffscalarization, steuer1983Tchebycheffscalarization, choo1983Tchebycheffefficiency}. However, the min-max formulation is non-differentiable and can be challenging to optimize in practice \citep{goffin1977convergence}. \citet{lin2024smooth} introduce smooth Tchebycheff scalarization (STS), which instead optimizes $\min_z \tau \log \sum_{i=1}^{k} \exp(\lambda_i(L_i(z) - L_i^\star) / \tau)$. Here, $\tau$ is a smoothing parameter, and STS approaches TS as $\tau \to 0^+$. STS has many of the same favorable properties as classical TS while being more amenable to gradient-based optimization due to its smoothness.

\section{Multi-Objective Offline Reinforcement Learning}
In multi-objective (KL-constrained or MaxEnt) RL, we wish to simultaneously optimize $k$ reward functions $r_1, \ldots, r_k$. Letting $\eta = 1/(2\gamma^2)$, this is equivalent to the optimization

\setlength{\abovedisplayskip}{-11pt}
\begin{align}
    &\argmax_{\pi} \E_{\substack{x \sim \mathcal{P} \\ y \sim \pi(\cdot \mid x)}} \left[r_i(x, y) \right]_{i=1}^{k} \text{\quad s.t. \quad} \mathbb{D}_{\mathrm{KL}}(\pi \mid\mid \pi_0) \le 1 / (2\gamma^2) \label{eq:kl-constrained} \\
    =& \argmin_{\pi} \E_{\substack{x \sim \mathcal{P} \\ y \sim \pi(\cdot \mid x)}} \left[\log \pi(y | x) - \log \pi_i^\star(y | x) \right]_{i=1}^{k}
    \label{eq:morl} \\
    \pi_i^{\star}(y | x) =& \frac{1}{Z_i(x)} \pi_0(y | x) \exp\left(\frac{r_i(x, y)}{\gamma \sqrt{\mathbb{V}_{\pi_0}[r_i(x, y)]}} \right) \label{eq:pi_i_star}
\end{align}
\setlength{\abovedisplayskip}{\belowdisplayskip}In practice, we estimate $\sigma_i^2 = \mathbb{V}_{\pi_0}[r_i(x, y)]$ as the empirical variance of $r_i(x, y)$ on the full training set. Most recent MORL work linearly scalarizes the reward to reduce the problem to single-objective RL \citep{li2021morl, zhou2024modpo, wen2025paretooptimal, chennakesavalu2025aligning, yang2019morl}. It is straightforward to show that this is equivalent to linearly scalarizing Optimization Problem~\ref{eq:morl}. Others consider alternative reward scalarizations including STS, but they do not directly correspond to alternative scalarizations of Optimization Problem~\ref{eq:morl} \citep{moffaert2013scalarized, williams2024morl, roijers2013survey, hayes2022survey, mccarter2026scsguidance, qiu2024traversingparetooptimalpolicies}.

\subsection{Smooth Tchebycheff Reward Scalarization}
\label{sec:scrs}
We innovate on prior work by applying smooth Tchebycheff scalarization (STS) \citep{lin2024smooth} directly to Optimization Problem~\ref{eq:morl}. Our goal is to derive a scalarized reward $R$ with the property that an $R$-optimal policy simultaneously minimizes an upper bound on its KL divergence from multiple optimal policies $\pi_1^\star, \ldots, \pi_k^\star$, one for each reward $r_1, \ldots, r_k$:
\begin{align}
    & \min_\pi \tau \log \sum_{i=1}^{k} \exp\bigg( \frac{\lambda_i}{\tau} \E_{\substack{x \sim \mathcal{P} \\ y \sim \pi(\cdot \mid x)}} \left[\log \pi(y | x) - \log \pi_i^\star(y | x) \right] \bigg) \notag \\
    \le& \min_\pi \E_{\substack{x \sim \mathcal{P} \\ y \sim \pi(\cdot \mid x)}} \left[ \tau \log \sum_{i=1}^{k} \exp\left( \frac{\lambda_i}{\tau} (\log \pi(y | x) - \log \pi_i^\star(y | x)) \right)\right] \notag \\
    =& \min_\pi \lambda_{\min} \E_{\substack{x \sim \mathcal{P} \\ y \sim \pi(\cdot \mid x)}} \left[ \log \pi(y | x) -  \frac{\tau}{\lambda_{\min}} \log \left( \sum_{i=1}^{k} \frac{\pi(y | x)^{(\lambda_i - \lambda_{\min})/\tau}}{\pi_i^\star(y | x)^{\lambda_i/\tau}} \right)^{-1} \right] \label{eq:optimization_problem} \\
    \le& \min_\pi \lambda_{\min} \E_{\substack{x \sim \mathcal{P} \\ y \sim \pi(\cdot \mid x)}} \left[ \log \pi(y | x) -  \frac{\tau}{\lambda_{\min}} \log \left(\sum_{i=1}^{k} \frac{1}{\pi_i^\star(y | x)^{\lambda_i/\tau}} \right)^{-1} \right] \label{eq:optimization_problem_relax}
\end{align}
The first upper bound is due to Jensen's inequality, while the second follows from the fact that $\pi(y | x)^{(\lambda_i - \lambda_{\min})/\tau} \le 1$, with equality when $\lambda_1 = \cdots = \lambda_k = 1/k$. To avoid problematic dependencies between the scalarized reward and a potentially sub-optimal reference $\pi_0$, we let $\pi_i^{\star}(y | x) \propto \exp(r_i(x, y) / (\gamma \sigma_i))$ be the optimal policy under MaxEntRL.\footnote{We also ran initial experiments with a variant of this reward scalarization derived from KL-constrained RL, i.e. $\pi_i^\star(y | x) \propto \pi_0(y | x) \exp(r_i(x, y) / (\gamma \sigma_i))$. It significantly underperformed $\pi_0$-independent reward scalarization because $\pi_0$ introduces high variance when estimating $Z_i(x)$.} Then, Optimization Problem~\ref{eq:optimization_problem} is a KL minimization that is equivalent to MaxEntRL with reward
\begin{align}
    R_{\text{ST}}^{\lambda,\pi}(x, y; \gamma) \! = -\frac{\tau \gamma}{\lambda_{\min}}\! \log \sum_{i=1}^{k} \exp\!\left(\frac{\lambda_i \!-\! \lambda_{\min}}{\tau} \log \pi(y | x) - \frac{\lambda_i}{\tau} \bigg( \frac{r_i(x, y)}{\gamma \sigma_i} - \log Z_i(x) \bigg) \! \right) \label{eq:reward}
\end{align}
Similarly, the relaxed Optimization Problem~\ref{eq:optimization_problem_relax} is equivalent to MaxEntRL with reward
\begin{align}
    R_{\text{ST}}^{\lambda}(x, y; \gamma) = -\frac{\tau \gamma}{\lambda_{\min}} \log \sum_{i=1}^{k} \exp\left( -\frac{\lambda_i}{\tau} \bigg( \frac{r_i(x, y)}{\gamma \sigma_i} - \log Z_i(x) \bigg) \right) \label{eq:reward_relax}
\end{align}
The policy-independent $R_{\text{ST}}^{\lambda}$ is useful in settings where the dependency of $R_{\text{ST}}^{\lambda,\pi}$ on the training policy $\pi$ could cause training instability.
In practice, we use the training set $\{(x_m, y_m^{(n)}): m \in [M], n \in [N_m]\}$ to estimate  $\log \hat{Z}_i(x_m) = \log \sum_{n = 1}^{N_m} \exp\big( r_i(x_m, y_m^{(n)})/(\gamma \sigma_i) \big)$.

This estimator ensures the {\em distribution-relative reward} $\rho_i(x, y) = r_i(x, y) / \sigma_i - \gamma \log \hat{Z}_i(x) \le 0$, a necessary condition for the theoretical guarantees of \citet{lin2024smooth}. \citet{qiu2024traversingparetooptimalpolicies} instead apply STS directly to the rewards, yielding $\rho'_i(x, y) = (r_i(x, y) - \iota_i - \max_y r_i(x, y)) / \kappa_i$, where the offsets $\iota_i$ and scales $\kappa_i$ are free hyperparameters. Our derivation provides optimal $\iota_i = \gamma \sigma_i \log \hat{Z}_i(x) - \max_y r_i(x, y)$ and $\kappa_i = \sigma_i$ in terms of a single hyperparameter $\gamma$.

\begin{figure}[tbp]
    \centering
    \includegraphics[width=\linewidth, trim={0.25cm 0.3cm 0.25cm 0.25cm}, clip]{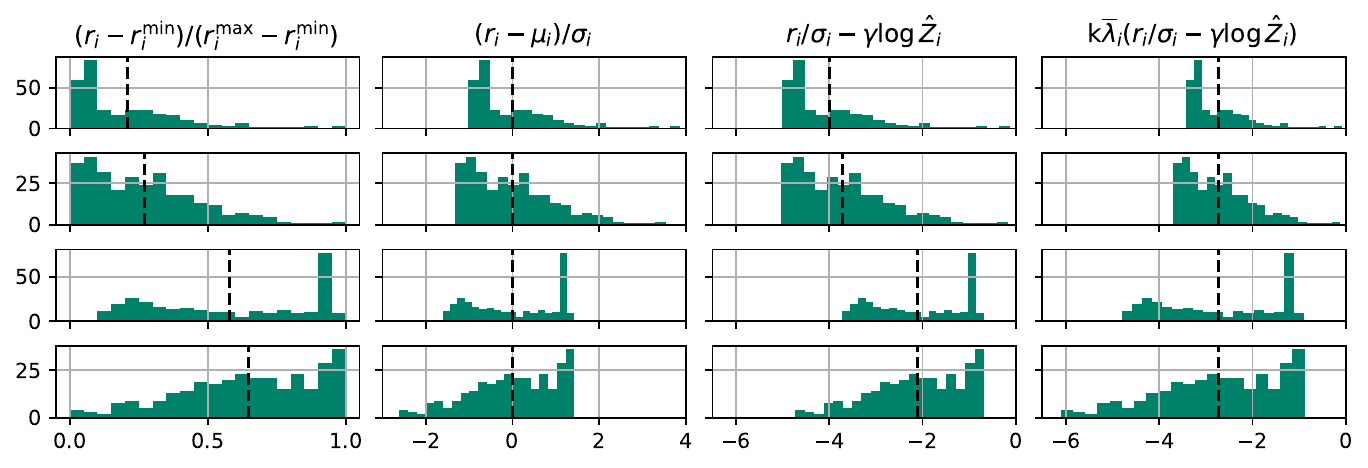}
    \caption{Visualizing different reward standardizations. The vertical dashed line indicates the mean of each distribution. Column 1: min-max standardization. Column 2: mean-variance standardization. Column 3: distribution-relative reward $\rho_i(x, y)$. Column 4: $\rho_i(x, y)$ reweighted by $\bar{\lambda}$ (Equation~\ref{eq:lambda_bar}) so all $\bar{\lambda}_i \rho_i(x, y)$ have the same mean.}
    \label{fig:standardizations}
    \vspace{-11pt}
\end{figure}

To build intuition for Equation~\ref{eq:reward_relax}, we note that $\lim_{\tau \to 0^+} \lambda_{\min} R_{\text{ST}}^{\lambda}(x, y; \gamma) = \min_i \lambda_i \rho_i(x, y)$. So as $\tau \to 0^+$, sequences $y$ with high $R_{\text{ST}}^{\lambda}(x, y; \gamma)$ must simultaneously have high distribution-relative rewards $\rho_i(x, y)$ for all $i$. We now visualize different standardizations for 4 rewards in Figure~\ref{fig:standardizations}. Rewards $r_{1-2}$ are left-skewed, while $r_{3-4}$ are right-skewed. 

A good standardization should penalize the rare sequences with low $r_{3-4}$ more than the common sequences with low $r_{1-2}$. Conversely, it should favor high $r_{1-2}$ more than high $r_{3-4}$. Min-max (Column 1) does the opposite. Mean-variance (Column 2) achieves this goal to some extent, but the effect is limited by design and over-indexes on outliers. The $Z_i$'s let the distribution-relative rewards $\rho_i$ (Column 3) favor high $r_{1-2}$ more than high $r_{3-4}$ without placing as much weight on outliers as mean-variance standardization. Now, let
\begin{align}
    \bar{\lambda}_i \propto \left( \frac{1}{M}\sum_{m=1}^{M} \frac{1}{N_m} \sum_{n=1}^{N_m} -\rho_i(x_m, y_m^{(n)}) \right)^{-1}
    \label{eq:lambda_bar}
\end{align}
re-weight the $\rho_i$'s to all have the same mean. $\bar{\lambda}_i\rho_i(x, y)$ (Column 4) compresses left-skewed rewards $r_{1-2}$ and stretches right-skewed $r_{3-4}$ to penalize the rare sequences with low $r_{3-4}$ more than the common sequences with low $r_{1-2}$. $r_4$ is more right-skewed than $r_3$, so this effect is even more pronounced. Different preferences $\lambda$ achieve different tradeoffs.

\subsection{\methodfull{}}
\label{sec:scpo}
To design a practical offline MORL algorithm, we define the difference
\begin{align}
    \!\!D_{\text{ST}}^{\lambda, \pi}(x, y_w, y_l; \beta, \gamma) &= \beta \bigg(\!\! \log \frac{\pi(y_w | x)}{\pi_0(y_w | x)} \!-\! \log \frac{\pi(y_l | x)}{\pi_0(y_l | x)} \!\bigg) \!-\! \left(\! R_{\text{ST}}^{\lambda, \pi}(x, y_w; \gamma) \!-\! R_{\text{ST}}^{\lambda, \pi}(x, y_l; \gamma) \right) \label{eq:scpo_difference}
\end{align}
We then adapt OffsetDPO \citep{amini2024offsetdpo} (Equation~\ref{eq:odpo}) to get the \methodfull{} (\method{}) loss
\begin{align}
    \mathcal{L}_{\text{\method{}}}(\pi; \alpha, \beta, \gamma, \delta, \lambda) =
    \!\!\!\!\!\!\!\!\!\!\!\!\!\!\!\!\!\!\!\!\! 
    \sum_{\substack{x, y_w, y_l \\ R_{\text{ST}}^{\lambda}(x, y_w; \gamma) > R_{\text{ST}}^{\lambda}(x, y_l; \gamma) + \delta}}
    \!\!\!\!\!\!\!\!\!\!\!\!\!\!\!\!\!\!\!\!\!\!\!
    - \log \sigma \Big( D_{\text{ST}}^{\lambda, \pi}(x, y_w, y_l; \beta, \gamma) + \delta \Big) - \frac{\alpha}{|y_w|} \log \pi(y_w | x) \label{eq:scpo_loss}
\end{align}
Notably, we use the policy-independent reward $R_{\text{ST}}^{\lambda}$ (Equation~\ref{eq:reward_relax}) to determine preference pairs, since having preferences evolve with the training policy $\pi$ led to instability in early experiments. However, we use the policy-dependent $R_{\text{ST}}^{\lambda, \pi}$ (Equation~\ref{eq:reward}) in the individual loss terms because it arises from a tighter bound to the optimization problem we are solving.
We also include the length-averaged NLL of the winners in each comparison \citep{pang2024irpo}. This regularization is especially important when the training dataset is not sampled from the reference policy $\pi_0$, a common reality in offline RL \citep{fu2026srft, liu2024rpo}.

\section{Experiments}
\label{sec:experiments}
%While much of the PLM literature has focused on evaluating the correlation between a protein's biological attributes and the likelihood a PLM assigns it \citep{notin2023proteingym, meier2021esm1v, notin2023proteinnpt, zhao2024confit, bhatnagar2025progen3, sun2024aidoprotein, widatalla2024proteindpo, nijkamp2023progen2}, this approach is inappropriate for multi-objective optimization where individual attributes may be uncorrelated or even anti-correlated. Instead, we turn to the problem statement of RL and estimate the expected rewards of sequences sampled from the PLM. We rely on the hypervolume enclosed by the non-dominated solutions of the expected rewards as our primary performance indicator \citep{auger2009hypervolumeindicator, yang2019mobayesopt}.

\subsection{Methods}
\label{sec:methods}
As baselines, we test direct preference optimization (DPO \citep{rafailov2023dpo}, Equation~\ref{eq:dpo}) and OffsetDPO (ODPO \citep{amini2024offsetdpo}, Equation~\ref{eq:odpo}) using linear reward scalarization $\sum_{i=1}^{k} \lambda_i r_i(x, y) / \sigma_i$. We refer to these methods as DPO-Lin and ODPO-Lin (identical to Pareto-Optimal Energy Alignment by \cite{wen2025paretooptimal}). Multi-objective DPO (MODPO \citep{zhou2024modpo}) is a popular method similar to ODPO-Lin, but it defines preferences $y_w \succ y_l$ by a {\em specific} reward $r_j$ and uses the remaining $r_{-j}$ for the margin in the loss
\begin{align*}
    \sum_{\substack{x, y_w, y_l \\ r_j(x, y_w) > r_j(x, y_l) + \delta}}
    \!\!\!\!\!\!\!\!\!\!\!\!\!\!\!
    -\log \sigma \Bigg( \frac{\beta}{\lambda_j} \left(\log \frac{\pi(y_w | x)}{\pi_0(y_w|x)} - \log \frac{\pi(y_l | x)}{\pi_0(y_l|x)} \right) - \sum_{i \neq j} \frac{\lambda_i}{\lambda_j} \left(\frac{r_i(x, y_w) - r_i(x, y_l)}{\sigma_i} \right) \Bigg)
\end{align*}
Where DPO-style algorithms are margin maximizers, Energy Rank Alignment (ERA, \citep{chennakesavalu2025aligning}, Appendix~\ref{app:era}) instead matches the aligned model's distribution to a specific target and has been used for protein optimization \citep{ibarraran2026efficient}. In the DPO family of methods, we also evaluate ODPO-STZ, i.e.\ ODPO with the smooth Tchebycheff-like z-score (STZ) scalarization proposed by \citet{mccarter2026scsguidance}:
\begin{align*}
    R_{\text{STZ}}^{\lambda}(x, y) = -\tau \gamma \log \sum_{i=1}^{k} \exp\left( - \frac{\lambda_i}{\tau \gamma} \frac{r_i(x, y) - \mu_i}{\sigma_i} \right)
\end{align*}
Finally, we evaluate our method \method{} (Equation~\ref{eq:scpo_loss}), which centers each reward $r_i / \sigma_i$ at $\gamma \log \hat{Z}_i(x)$ instead of its mean. STZ suffers from pathological behavior when reweighting the rewards by $\lambda$ because the terms inside the logsumexp are both positive and negative \citep{lin2024smooth}. \method{} avoids this issue because $r_i / \sigma_i - \gamma \log \hat{Z}_i(x) \le 0$ uniformly.

For each (dataset, base model), we train all algorithms with a range of preferences $\lambda$ to probe their Pareto fronts.\footnote{We actually train \method{} with the preference vector $\lambda_i' = \lambda_i \bar{\lambda}_i / \sum_{i=1}^{k} \lambda_i \bar{\lambda}_i$. Recall that $\bar{\lambda}$ (Equation~\ref{eq:lambda_bar}) re-weights the individual distribution-relative rewards to have the same mean. Thus, the transformation $\lambda \mapsto \lambda'$ simply keeps the preference vector interpretable for \method{}.} We aggregate over $\lambda$ and report the expected reward hypervolume of sequences sampled from aligned models \citep{auger2009hypervolumeindicator, yang2019mobayesopt}. We train 3 replicates of each configuration with different random seeds and dataset orders.

\subsection{Protein Engineering}
\label{sec:protein-evals}
\paragraph{Datasets} First, we maximize DHFR activity both in the absence and in the presence of an antibiotic, two uncorrelated objectives (Spearman $\rho = -0.010$) \citep{romanowicz2025dhfr}. The train and test splits comprise 310 and 139 diverse homologs.
Next, we maximize the negatively correlated activity and specificity of PbrR (Spearman $\rho = -0.806$) \citep{wang2025activelearningde}. We train on 1076 single-mutation variants and 76 variants with 2-5 mutations and test on 869 engineered variants with up to 12 mutations.
Finally, we maximize the activity, expression, and stability of \textalpha-amylase (Spearman $\rho = 0.655, 0.602, 0.719$ for A/E, A/S, E/S) \citep{armer2025protein-eng-tournament}. We train on 8076 single-mutation variants and test on 1879 variants with 4-10 mutations. These are challenging splits that test model generalization.

\paragraph{Models} We align protein language models (PLMs) on laboratory data measuring a variety of biological attributes to be optimized. We apply all methods to the autoregressive ProGen3-3B (PG3) \citep{bhatnagar2025progen3} and two internal retrieval-augmented PLMs similar to PoET \citep{truong2023poet}: ProGen-RA-3B (RA-3B) and ProGen-RA-10B (RA-10B) with 3B and 10B parameters respectively. See Appendix~\ref{app:protein-hyperparams} for additional details.

\setlength{\tabcolsep}{3.5pt}
\begin{table}[tpb]
    \centering
    \fontsize{8}{10}\selectfont
    \begin{tabular}{|ll|rrrrrr|}
    \hline
    Dataset & Model & DPO-Lin & ODPO-Lin & MODPO & ERA & ODPO-STZ & \method{} \\
    \hline
    DHFR & PG3 & \textbf{17.8 $\pm$ 0.4} & \textbf{17.7 $\pm$ 0.4} & \textbf{18.2 $\pm$ 0.1} & 13.9 $\pm$ 0.1 & \textbf{16.9 $\pm$ 1.3} & \textbf{18.2 $\pm$ 0.1} \\
    DHFR & RA-3B & 17.1 $\pm$ 0.3 & 16.4 $\pm$ 0.6 & 17.2 $\pm$ 0.2 & 16.7 $\pm$ 0.1 & 17.2 $\pm$ 0.1 & \textbf{17.9 $\pm$ 0.1} \\
    DHFR & RA-10B & \textbf{18.1 $\pm$ 0.1} & \textbf{17.8 $\pm$ 0.1} & \textbf{18.2 $\pm$ 0.1} & 14.8 $\pm$ 0.1 & 17.0 $\pm$ 0.2 & \textbf{18.0 $\pm$ 0.1} \\
    \hline
    PbrR & PG3 & 15.7 $\pm$ 0.5 & 15.0 $\pm$ 0.4 & 9.7 $\pm$ 0.7 & 13.8 $\pm$ 0.1 & 14.8 $\pm$ 0.2 & \textbf{17.4 $\pm$ 0.1} \\
    PbrR & RA-3B & 13.9 $\pm$ 0.5 & 15.3 $\pm$ 0.3 & 9.4 $\pm$ 0.4 & 15.9 $\pm$ 0.4 & 12.9 $\pm$ 0.2 & \textbf{17.9 $\pm$ 0.2} \\
    PbrR & RA-10B & 17.2 $\pm$ 0.1 & 17.1 $\pm$ 0.3 & 12.3 $\pm$ 0.3 & 11.4 $\pm$ 0.1 & 17.3 $\pm$ 0.1 & \textbf{17.8 $\pm$ 0.1} \\
    \hline
    \textalpha-Amylase & PG3 & 22.9 $\pm$ 0.1 & 22.9 $\pm$ 0.1 & 23.2 $\pm$ 0.1 & 19.6 $\pm$ 0.1 & 22.8 $\pm$ 0.1 & \textbf{27.8 $\pm$ 0.9} \\
    \textalpha-Amylase & RA-3B & 9.5 $\pm$ 0.3 & \textbf{12.2 $\pm$ 0.8} & 8.0 $\pm$ 0.2 & 1.4 $\pm$ 0.1 & 9.4 $\pm$ 0.3 & \textbf{11.3 $\pm$ 0.4} \\
    \textalpha-Amylase & RA-10B & \textbf{19.3 $\pm$ 0.1} & 17.1 $\pm$ 0.1 & 15.4 $\pm$ 0.3 & 4.7 $\pm$ 0.5 & 18.6 $\pm$ 0.2 & \textbf{19.5 $\pm$ 0.1} \\
    \hline
    \end{tabular}
    \caption{Hypervolumes (with standard errors) of offline off-policy expected rewards in the protein setting. \textbf{Bold} entries are indistinguishable from the best method with $p < 0.05$ in a 1-sided $t$-test. \method{} achieves or ties for the highest hypervolume in all settings.}
    \label{tab:offline-hypervolumes}
    \vspace{-11pt}
\end{table}

\paragraph{Offline Off-Policy Evaluations} In offline RL where we have access to a test set but cannot easily evaluate generations from the trained policy $\pi$, accurately estimating the expected reward $\E_{\pi}[r_i(x, y)]$ is a common challenge. We estimate expected rewards using the weighted importance sampling (WIS) estimator on the test set $\{(x_m, y_m^{(n)}): m \in [M], n \in [N_m]\}$ \citep{suttonbarto1998rl, precup2001offpolicy, practicalqmc, mahmood2014wis}:
\begin{gather*}
    \frac{1}{M} \sum_{m=1}^{M} \frac{1}{N_m} \sum_{n=1}^{N_m} w_m^{(n)} r_i(x_m, y_m^{(n)}) \text{\quad where \quad} w_m^{(n)} = \frac{\pi(y_m^{(n)} | x_m) / \pi_0(y_m^{(n)} | x_m) }{\sum_{n'=1}^{N_m} \pi(y_m^{(n')} | x_m) / \pi_0(y_m^{(n')} | x_m)}
\end{gather*}
While WIS is a biased estimator, it has much lower variance than vanilla importance sampling. We use WIS to estimate the expected rewards of every checkpoint from \{20\%, 30\%, \ldots, 90\%, 100\%\} of training. We then identify the resulting Pareto fronts across all preference vectors used and report their hypervolumes (Appendix~\ref{app:hv-estimator}) in Table~\ref{tab:offline-hypervolumes}. We also visualize example Pareto fronts and present a more qualitative analysis in Appendix~\ref{app:pareto-fronts}.

Most methods achieve near-identical performance for the DHFR dataset. This stems from a limitation of the offline evaluation: this dataset comprises diverse homologs that the models can easily distinguish, and all methods except ERA consistently assign high likelihoods to the few test sequences with high hypervolumes.

Conversely, the PbrR and \textalpha-amylase datasets are local mutational landscapes. Thus, they represent much denser samples of protein space that provide more reliable estimates of expected rewards via importance sampling. \method{} achieves the highest expected hypervolume in all 6 of these settings and is the sole winner in 4/6. Meanwhile, there are no consistent winners between the baselines. MODPO and ERA consistently underperform. MODPO's poor performance is explained by the fact that its ranking reward $r_j$ is poorly correlated with its margin reward(s) $r_{-j}$, especially for PbrR (Spearman $\rho = -0.806)$.

\paragraph{Generative Evaluations} For each (dataset, base model, algorithm) combination, we generate large batches of novel sequences from the two checkpoints that maximize offline expected hypervolume. See Appendix~\ref{app:plm-generations} for more details.
We then train Gaussian process (GP) regressors to estimate the expected rewards of these generations. GPs have proven viable protein property predictors in challenging interpolative and extrapolative settings \citep{groth2024kermut,truong2025poet2}, warranting their use as oracles. 

\begin{figure}[t]
    \centering
    \includegraphics[width=0.98\linewidth, trim={0.25cm 0.25cm 0.25cm 0.25cm}, clip]{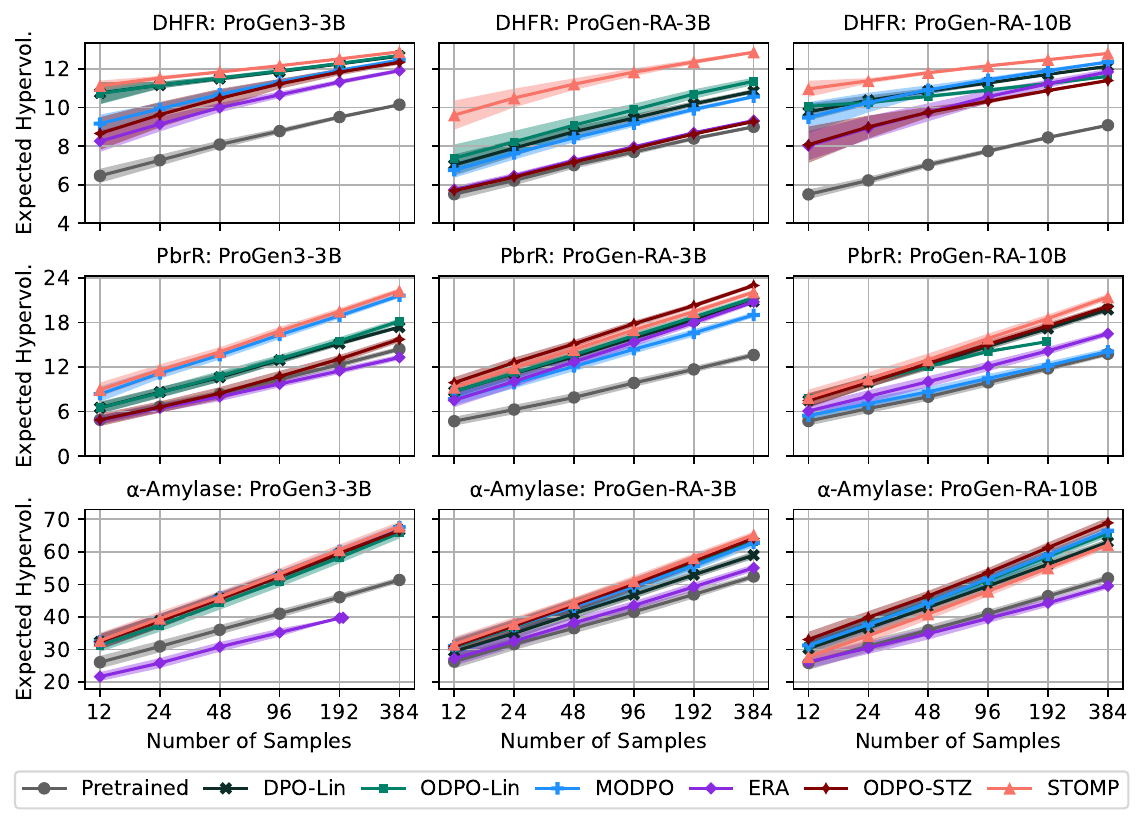}
    \caption{Hypervolumes of $k$ randomly sampled proteins from models aligned using different preference optimization algorithms. Error bands correspond to one standard error. \method{}'s generations achieve or tie for the highest expected hypervolumes in 7/9 settings.}
    \label{fig:gen-evals}
    \vspace{-11pt}
\end{figure}

In Figure~\ref{fig:gen-evals}, we compute the expected hypervolumes (along with their uncertainties) achieved by random samples of $k \in \{12, 24, 48, 96, 192, 384\}$ generated sequences. These represent realistic sample sizes that one might test in a small wet lab experiment. See Appendix~\ref{app:gp-reward-models} for more details on the reward models and hypervolume estimation.

Unlike the offline evaluation, clear differences emerge when we evaluate models on the more challenging DHFR generation task (Row 1). All base models aligned with \method{} generate sequences with the best predicted rewards. For PbrR (Row 2), \method{} achieves the highest expected hypervolume for PG3 and RA-10B, and the second-highest for RA-3B where it loses by a small margin to ODPO-STZ. For \textalpha-amylase (Row 3), most methods perform quite similarly to each other. This likely arises from a limitation of the GP reward models, as their predictive accuracy is lower on this dataset than the others (Table~\ref{tab:surrogate-evals}).

As in the off-policy evaluations, ERA is the worst method, while DPO-Lin and ODPO-Lin are in the middle of the pack. MODPO is one of the worst performers for DHFR with PG3 and RA-3B, as well as PbrR with the RA models, but it is one of the best for DHFR with RA-10B and PbrR with PG3. This contrasts with the off-policy evaluations, where it underperformed across the board. Similarly, ODPO-STZ is the best method for PbrR with both RA models but one of the worst for DHFR with all models and PbrR with PG3.

\subsection{Chatbot Alignment}
\label{sec:llm_benchmarks}
\paragraph{Datasets} First, we maximize helpfulness (\texttt{PKU-Alignment/beaver-7b-v1.0-reward}) and harmlessness (negative \texttt{PKU-Alignment/beaver-7b-v1.0-cost}) for the SafeRLHF dataset (Spearman $\rho = -0.445$) \citep{dai2024saferlhf}.
Next, we maximize helpfulness (\texttt{Ray2333/gpt2-large-helpful-reward\_model}) and harmlessness (\texttt{Ray2333/gpt2-large-helpful-reward\_model}) for AnthropicHH (Spearman $\rho = -0.614$) \citep{bai2022anthropichh, yang2024anthropichhrewards}. Finally, we maximize helpfulness, complexity, and negative verbosity for HelpSteer2 \citep{wang2024helpsteer2} (Spearman $\rho = 0.289, -0.348, -0.594$ respectively for helpfulness/complexity, helpfulness/--verbosity, and complexity/--verbosity), with \texttt{RLHFlow/ArmoRM-Llama3-8B-v0.1} heads as reward models \citep{wang2024armorm, wang2024dpa}.

\paragraph{Models} We train \texttt{Llama-3.1-8B-Instruct} \citep{grattafiori2024llama3} and \texttt{Qwen-2.5-7B-Instruct} \citep{qwen2025qwen25} on chatbot alignment datasets. We apply supervised fine-tuning using a subset of each train split and generate 8 novel completions for all remaining training prompts to curate an alignment dataset. See Appendix~\ref{app:llm-hyperparams} for additional details.

\paragraph{Generative Evaluations} We compare the same methods as in the PLM setting, but at matched KL divergence drift from the reference policy, as stated in the original Optimization Problem~\ref{eq:kl-constrained}. While all methods had similar KL drift in the PLM setting, the explicit constraint was necessary here for fair comparisons because we observed significantly different KL drift between runs in this setting \citep{rafailov2023dpo, rafailov2024daascaling, beirami2025bestofn}.

For every training checkpoint, we generate one response $y_x \sim \pi(\cdot | x)$ per test prompt $x$. We then estimate $\hat{\mathbb{D}}_{\mathrm{KL}}(\pi \mid\mid \pi_0) = \frac{1}{|\mathcal{P}|} \sum_{x \in \mathcal{P}} \log \pi(y_x| x) - \log \pi_0(y_x| x)$. We set a KL budget $\eta$ as the largest $\hat{\mathbb{D}}_\mathrm{KL}$ achieved by {\em all} training runs for a dataset and select one checkpoint per run satisfying $\eta - 2 \le\hat{\mathbb{D}}_\mathrm{KL} \le \eta + 2$. Finally, we generate one completion per checkpoint per test prompt, aggregate over preference vectors $\lambda$ to construct a Pareto front, and report hypervolumes in Table~\ref{tab:llm-hypervolumes}. See Appendices~\ref{app:llm-generations}, \ref{app:kl}, and \ref{app:hv-estimator} respectively for more details on sequence generation, KL drift matching, and hypervolume estimation.

In 5/6 evaluation settings, \method{}'s generations achieve or tie for the highest expected hypervolumes. In contrast to the protein experiments where its performance was more mixed, the simplest method DPO-Lin is the strongest performer amongst our baselines. It achieves or ties for the highest hypervolume in 3/6 settings, and it is second place to \method{} on HelpSteer2 with LLaMA. As in the protein evaluations, ERA is the worst method, and MODPO consistently underperforms DPO-Lin and OPDO-Lin. ODPO-STZ once again has highly inconsistent performance: it is in the bottom half of methods in 5/6 settings, but is tied with \method{} for the best algorithm on AnthropicHH with LLaMA.

\setlength{\tabcolsep}{3.5pt}
\begin{table}[t]
    \centering
    \fontsize{8}{10}\selectfont
    \begin{tabular}{|ll|rrrrrr|}
    \hline
    Dataset & Model & DPO-Lin & ODPO-Lin & MODPO & ERA & ODPO-STZ & \method{} \\
    \hline
    SafeRLHF & LLaMA & \textbf{13.0 $\pm$ 0.1} & 12.8 $\pm$ 0.3 & 11.3 $\pm$ 0.1 & 11.7 $\pm$ 0.2 & 12.1 $\pm$ 0.1 & \textbf{13.2 $\pm$ 0.1}  \\
    SafeRLHF & Qwen & 11.8 $\pm$ 0.1 & \textbf{12.1 $\pm$ 0.1} & 10.7 $\pm$ 0.1 & 6.2 $\pm$ 0.2 & 11.2 $\pm$ 0.2 & \textbf{12.2 $\pm$ 0.1} \\
    \hline
    AnthropicHH & LLaMA & 20.4 $\pm$ 0.1 & 20.8 $\pm$ 0.3 & 20.0 $\pm$ 0.1 & 17.3 $\pm$ 0.1 & \textbf{21.3 $\pm$ 0.2} & \textbf{21.1 $\pm$ 0.2} \\
    AnthropicHH & Qwen & \textbf{21.3 $\pm$ 0.1} & \textbf{21.1 $\pm$ 0.2} & 19.6 $\pm$ 0.1 & 15.5 $\pm$ 0.1 & 20.4 $\pm$ 0.1 & \textbf{21.3 $\pm$ 0.1} \\
    \hline
    HelpSteer2 & LLaMA & 81.2 $\pm$ 0.9 & 80.8 $\pm$ 0.7 & 71.4 $\pm$ 0.4 & 53.8 $\pm$ 0.5 & 67.6 $\pm$ 0.8 & \textbf{83.5 $\pm$ 0.6} \\
    HelpSteer2 & Qwen & \textbf{83.1 $\pm$ 0.5} & 80.2 $\pm$ 1.5 & 76.6 $\pm$ 0.8 & 49.5 $\pm$ 0.5 & 67.1 $\pm$ 0.4 & 78.8 $\pm$ 0.6 \\
    \hline
    \end{tabular}
    \caption{Hypervolumes (with standard errors) of generations from aligned LLMs. \textbf{Bold} entries are indistinguishable from the best method with $p < 0.05$ in a 1-sided $t$-test. \method{} achieves or ties for the highest hypervolume in 5/6 settings.}
    \label{tab:llm-hypervolumes}
    \vspace{-11pt}
\end{table}

\section{Discussion}
\label{sec:discussion}
In multi-objective reinforcement learning (MORL), a fundamental limitation of linear reward scalarization is its inability to recover the full Pareto front of optimal solutions.
Smooth Tchebycheff scalarization (STS) is a recent technique that overcomes this limitation. Rather than scalarizing the rewards, we frame MORL itself as the optimization problem to be scalarized via STS and use this insight to propose \methodfull{} (\method{}), a novel offline MORL algorithm.

Our experiments span multiple language models and diverse datasets in both the protein engineering and chatbot alignment domains. They demonstrate that \method{} outperforms existing baselines, achieving or tying for the highest hypervolumes in 16/18 protein settings and 5/6 natural language settings. Other algorithms rank inconsistently relative to each other in the protein experiments, while vanilla DPO with linear scalarization is a strong baseline for the natural language experiments. ODPO-STZ relies on a more naive application of STS than \method{} and shows particularly inconsistent performance. This highlights the importance of our principled derivation in Section~\ref{sec:scrs} in elucidating the appropriate techniques for applying STS to offline MORL.

More broadly, the smooth Tchebycheff reward scalarizations derived from Optimization Problems~\ref{eq:optimization_problem}--\ref{eq:optimization_problem_relax} are completely general to any KL-constrained or MaxEnt RL formulation. A natural extension would be applying these scalarizations to multi-objective {\em online} RL. A key challenge to overcome in this setting would be estimating the partition functions $Z_i(x)$ when sequences $y$ are not collated in an offline dataset but are instead generated online. 

Finally, we leveraged the framing of MaxEntRL to scalarize the rewards $R_{\text{ST}}^{\lambda, \pi}$ and $R_{\text{ST}}^{\lambda}$ (Equations~\ref{eq:reward}--\ref{eq:reward_relax}) because $\pi_0$ introduced high variance when estimating partition functions. However, \method{} implicitly constrains the KL divergence between the $\pi$ and $\pi_0$. This is because our early experiments found that MaxEnt variants of \method{} (i.e.\ removing the $\log \pi_0(y | x)$ terms from Equation~\ref{eq:scpo_difference}) suffered from training instability and sensitivity to hyperparameters. We leave further study of such reference-free losses to future work. 

\subsection*{Ethics Statmeent}
\label{app:safety}
Computational protein design carries the dual potential to accelerate the development of novel therapeutics and other society-improving molecules, while providing parallel capabilities for nefarious uses, such as engineering of bioweapons. When bolstered by current and future iterations of generative AI, these capabilities are heightened and expected to grow further. The global protein design community has begun to establish appropriate regulations and guidelines towards the continued beneficial development and application of these technologies. We support having a set of community values, guiding principles, and commitments for the responsible development of AI for protein design (\href{https://responsiblebiodesign.ai/}{https://responsiblebiodesign.ai}). Gene synthesis represents a critical step in the actualization of designed protein sequences. The International Gene Synthesis Consortium (IGSC) unites major gene synthesis providers under a commitment to screen all incoming orders against known pathogens and potentially dangerous sequences. As a concrete step towards safe application of protein design technology, all gene synthesis work in support of the present study was performed with IGSC members. For all protein design projects, we urge researchers to maintain ethical oversight throughout project initiation, experimental characterization, and subsequent deployment phases to ensure safety and avoid unintended harmful outcomes. For the current algorithms described in this paper, we find the benefit of research accessibility to greatly outweigh any theoretical risks.

\subsection*{AI Use Statement}
In this work, we used generative AI tools for the chatbot alignment experiments for scaffolding the codebase, as well as drafting data loading and metric aggregation for efficient experimental iteration.
We did not use generative AI tools for the writing of this manuscript, the derivation of \method{}, or the protein design experiments.
We have reviewed all AI-assisted work. This includes the authors reviewing LLM-generated code and carefully assessing the veracity of results produced with that code. We take responsibility for the final content of this work,
including text, claims or artifacts produced with the aid of generative AI.

\iftoggle{anonymous}{}{
\subsubsection*{Acknowledgments}
We would like to thank Joel Beazer, Sarthak Jain, and Nicholas Lucas-Randolph for helpful discussions and code reviews.

\subsubsection*{Competing Interests}
The authors are current or former employees, contractors, or executives of Profluent Bio, Inc. and may hold shares in Profluent Bio, Inc. Funding for this work was provided by Profluent Bio, Inc.
}

\bibliographystyle{abbrvnat}
{\footnotesize \bibliography{references}}

%%%%%%%%%%%%%%%%%%%%%%%%%%%%%%%%%%%%%%%%%%%%%%%%%%%%%%%%%%%%
\clearpage
\appendix

\section{Visualizing Pareto Fronts for Off-Policy Evaluations}
\label{app:pareto-fronts}
\begin{figure}[bh]
    \centering
    \includegraphics[width=0.95\linewidth, trim={0.25cm 0.25cm 0.25cm 0.25cm}, clip]{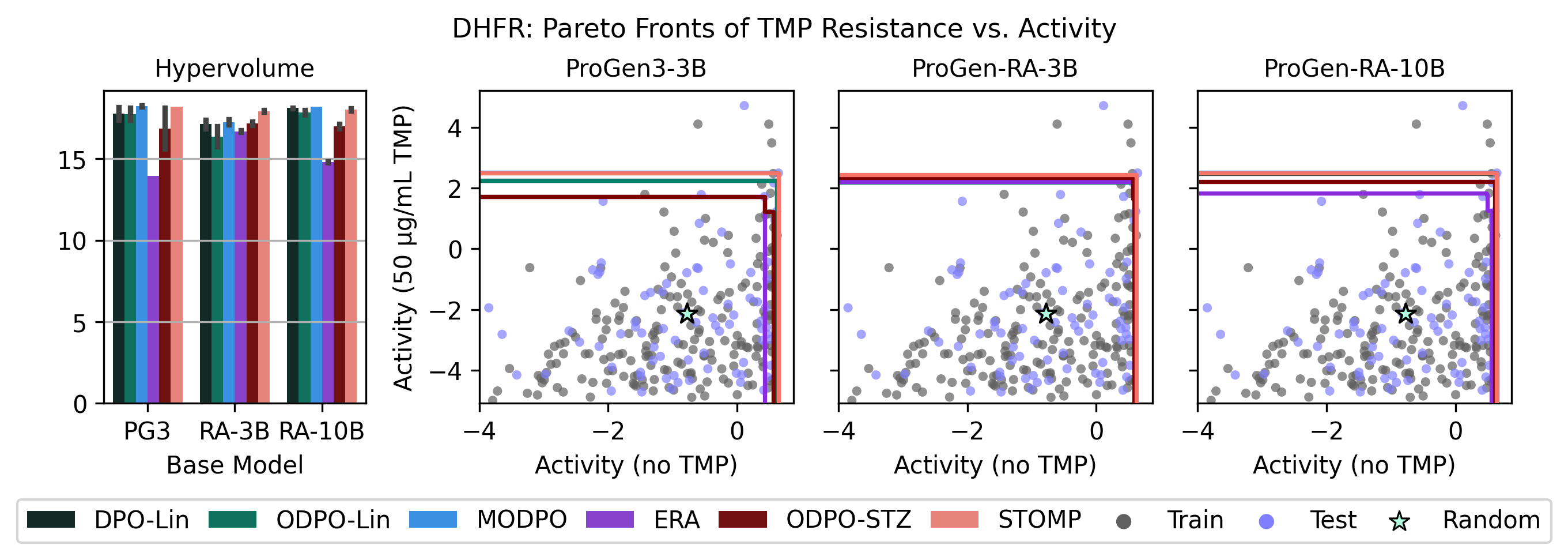}
    \caption{Pareto fronts of DHFR activity in the absence of TMP and in the presence of 50{\textmu}g/mL of TMP. All fitness values are in units of $\log_2$-fold change.}
    \label{fig:dhfr}
\end{figure}

\begin{figure}[bh]
    \centering
    \includegraphics[width=0.95\linewidth, trim={0.25cm 0 0.25cm 0}, clip]{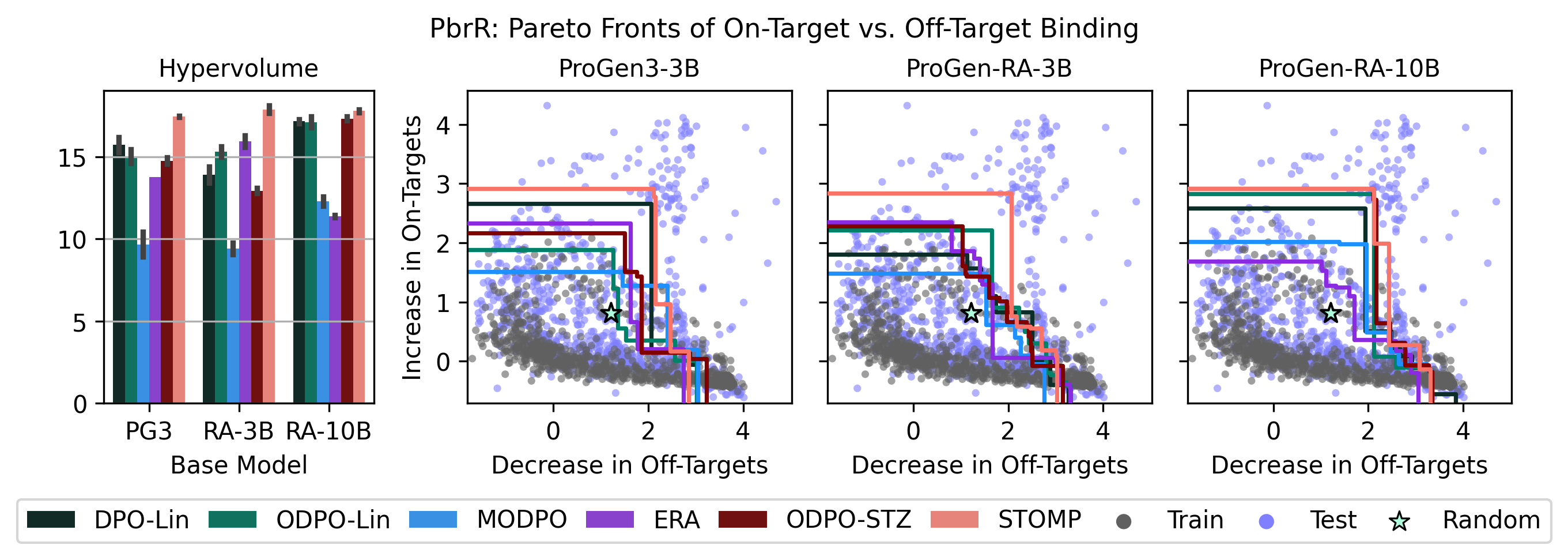}
    \caption{Pareto fronts of increase in on-target binding and decrease in off-target binding for PbrR. All fitness values are in units of $\log_2$-fold change.}
    \label{fig:pbrr}
\end{figure}

\begin{figure}[tb]
    \centering
    \includegraphics[width=\linewidth, trim={0.25cm 0.25cm 0.25cm 0.25cm}, clip]{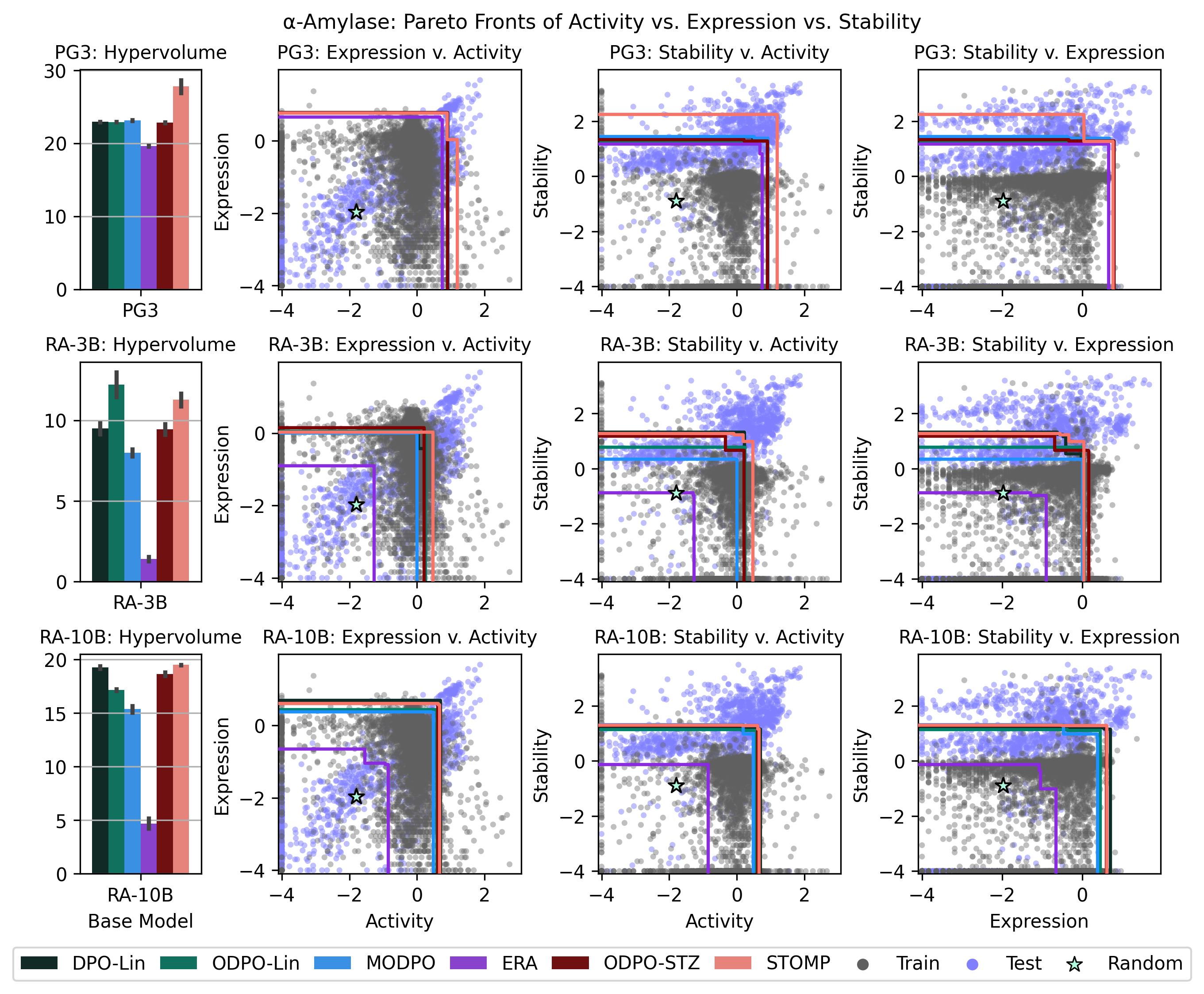}
    \caption{Pareto fronts of $\log_2$-fold change in \textalpha-amylase activity, stability, and expression.}
    \label{fig:amylase}
\end{figure}

In Figures~\ref{fig:dhfr}--\ref{fig:amylase}, we illustrate the Pareto fronts achieved by different methods in the offline off-policy evaluations described in Section~\ref{sec:protein-evals}. We visualize the training set in grey, the test set in lilac, and the expected value of a random point from the test set as a green star. Recall that we estimate $\E_{\pi}[r_i(x, y)]$ on the test set using the weighted importance sampling estimator \citep{suttonbarto1998rl, precup2001offpolicy, practicalqmc, mahmood2014wis}:
\begin{gather*}
    \frac{1}{M} \sum_{m=1}^{M} \frac{1}{N_m} \sum_{n=1}^{N_m} w_m^{(n)} r_i(x_m, y_m^{(n)}) \text{\quad where \quad} w_m^{(n)} = \frac{\pi(y_m^{(n)} | x_m) / \pi_0(y_m^{(n)} | x_m) }{\sum_{n'=1}^{N_m} \pi(y_m^{(n')} | x_m) / \pi_0(y_m^{(n')} | x_m)}
\end{gather*}

Using this estimator, the expected reward under $\pi_0$ is simply the expected reward of a random test point (the green star). Meanwhile, the expected reward under a preference optimized policy indicates that policy's most preferred sequences from the test set.

For DHFR (Figure~\ref{fig:dhfr}), almost all methods perform near-optimally. This is largely a consequence of the fact that the dataset comprises highly diverse homologs that all models can easily distinguish (Section~\ref{sec:protein-evals}). For PbrR (Figure~\ref{fig:pbrr}), where the two objectives are highly negatively correlated in the training set ($\rho = -0.806$), \method{} is the only method that consistently assigns high likelihood to the cluster of sequences exclusive to the test set that have both a high increase in on-target binding and a high decrease in off-target binding. Finally, for \textalpha-amylase (Figure~\ref{fig:amylase}), where the three objectives are positively correlated, \method{} slightly expands the Pareto front achieved by the other methods for PG3 and RA-10B, while achieving comparable performance to ODPO-Lin for RA-3B.

\section{Training Hyperparameters}
\label{app:hyperparams}

\subsection{Protein Engineering Experiments}
\label{app:protein-hyperparams}
Similar to \citet{bhatnagar2025progen3}, we use the \texttt{<bos>} token as the context $x$ for PG3. For the RA models, we follow PoET's prompt construction strategy \citep{truong2023poet} to construct 24 prompts $x_1, \ldots, x_{24}$ with up to 4096 tokens of homologs discovered via ColabFold search \citep{steinegger2017mmseqs, mirdita2022colabfold}. For the diverse DHFR dataset, we use homologs >30\% ID to any training sequence, while for the local mutational landscapes of PbrR and \textalpha-amylase, we use homologs >70\% ID to the wild type.

We train all models described in Section~\ref{sec:protein-evals} for 782 batches with a batch size of 64. We fine-tune all model weights as described by \citet[Appendix C]{bhatnagar2025progen3}.
We used the AdamW optimizer \citep{kingma2017adam,loshchilov2018adamw} with $\beta_1 = 0.9$, $\beta_2 = 0.95$, and BF16 mixed precision \citep{micikevicius2018mixedprecision}. We linearly increased the learning rate to $10^{-5}$ over an initial warmup period of 79 batches, after which we decay the learning rate to $5 \times 10^{-6}$ following a cosine schedule. All models were trained using PyTorch FSDP2 \citep{paszke2019pytorch, zhao2023fsdp} with gradient checkpointing \citep{chen2016gradcheckpoint} on 8xH200 for up to 3 hours.

We train each model on each dataset with 3 preference vectors $\lambda$. For the PbrR and DHFR datasets, we sweep over $\lambda \in \left\{ (\frac{1}{3}, \frac{2}{3}), (\frac{1}{2}, \frac{1}{2}), (\frac{2}{3}, \frac{1}{3}) \right\}$. For \textalpha-amylase, we use $\lambda \in \left\{ (\frac{1}{3}, \frac{1}{3}, \frac{1}{3}), (\frac{1}{4}, \frac{1}{4}, \frac{1}{2}), (\frac{2}{5}, \frac{1}{5}, \frac{2}{5}) \right\}$ ordered as (activity, expression, stability). We train 3 replicates of each configuration with different random seeds and dataset orders. In total, we train 486 distinct models.

\begin{table}[htbp]
    \centering
    \begin{tabular}{|ll|cccc|}
        \hline
        Dataset & Base Model & $\alpha$ & $\beta$ & $\gamma$ & $\delta$ \\
        \hline
        DHFR & ProGen3-3B & 0.05 & 0.2 & 0.2 & 1 \\
        DHFR & ProGen-RA3B & 0.01 & 0.2 & 0.2 & 1 \\
        DHFR & ProGen-RA10B & 0.05 & 0.2 & 0.2 & 1 \\
        \hline
        PbrR & ProGen3-3B & 0.02 & 0.1 & 0.2 & 1 \\
        PbrR & ProGen-RA3B & 0.02 & 0.1 & 0.2 & 1 \\
        PbrR & ProGen-RA10B & 0.01 & 0.1 & 0.2 & 1 \\
        \hline
        \textalpha-Amylase & ProGen3-3B & 0.05 & 0.05 & 0.2 & 0.5  \\
        \textalpha-Amylase & ProGen-RA3B & 0.02 & 0.05 & 0.2 & 0.5 \\
        \textalpha-Amylase & ProGen-RA10B & 0.02 & 0.05 & 0.2 & 0.5 \\
        \hline
    \end{tabular}
    \caption{Hyperparameters of preference optimization algorithms.}
    \label{tab:preference-opt-hyperparams}
\end{table}

Table~\ref{tab:preference-opt-hyperparams} details the hyperparameters of the preference optimization algorithms for each experimental setting. For each of the datasets and models, we use the same KL regularization strength $\beta$ and reward difference threshold $\delta$ across all algorithms after tuning the values for DPO-Lin. Similar to \citet{bhatnagar2025progen3}, we tune the NLL regularization coefficient $\alpha$ (Equation~\ref{eq:scpo_loss}) individually for each method to the lowest value ensuring that validation perplexity degrades by no more than 1 point over the course of training. In general, this results in consistent values of $\alpha$ (as reported in Table~\ref{tab:preference-opt-hyperparams}) for all preference optimization algorithms for each (dataset, base model) pair, with the following exceptions:
\begin{itemize}
    \item For PbrR, we set $\alpha = 0.05$ for ODPO-STZ and $\alpha = 0.01$ for \method{}. These settings are consistent across all models.
    \item For DHFR, we set $\alpha = 0.02$ for \method{} when using the ProGen3-3B and ProGen-RA-10B models.
    \item For \textalpha-amylase, we set $\alpha = 0.02$ for DPO-Lin when using the ProGen3-3B model.
\end{itemize}

We also improve the training stability of all methods by clamping the reward difference to 1, i.e. we take $\min(1, R(x, y_w) - R(x, y_l) + \delta)$. We fix \method{}'s and ODPO-STZ's scalarization temperatures to $\tau = 1$ and $\gamma = 0.2$ across the board because we find them to be good default values that work well in a wide range of settings (Appendix~\ref{app:gamma-tau}).

\subsection{Chatbot Alignment Experiments}
\label{app:llm-hyperparams}
For all three benchmark datasets, the pool of test prompts are taken from the published test set of each dataset. The training set for each benchmark is then split into three partitions: the SFT split, the alignment training split, and the alignment validation split. For each of the benchmarks, eight completions are sampled per-prompt in the alignment split. These are used for training rather than the existing responses from the dataset because the partition function estimation requires multiple samples per-prompt drawn from the same policy to construct an appropriate estimate, along with the population statistics required for reward standardization.

We train all models in Section~\ref{sec:llm_benchmarks} for 400 optimizer steps using an effective batch size of 16. All models are trained as LoRA adapters \citep{hu2022lora} with rank $r=16$, $\alpha=32$, dropout $0.05$, on all attention and feed-forward projections (\texttt{\{q,k,v,o,gate,up,down\}}\_\texttt{proj}). We use the AdamW optimizer \citep{kingma2017adam,loshchilov2018adamw} with $\beta_1 = 0.9$, $\beta_2 = 0.95$, and FP32 precision for the LoRA adapters. We linearly increase the learning rate to $10^{-5}$ over an initial warmup period of 40 optimizer steps, after which we decay the learning rate to $5 \times 10^{-6}$ following a cosine schedule. All models are trained on a single H200 until completion of the 400 optimizer steps.

We set alignment hyperparameters $\beta = 0.1$ and $\alpha = 0$ (no NLL regularization), with scalarization temperatures $\tau = 1.0$ and $\gamma = 0.2$ as in the protein experiments. We found that NLL regularization was not necessary in this setting because the training sets were sampled from the reference policy $\pi_0$ \citep{fu2026srft, liu2024rpo}. For each (dataset, base model) combination, we set $\delta$ such that 30\% of comparison pairs were retained across all objectives compared. The reference policy $\pi_0$ is a supervised fine-tune of the base model on the dataset's SFT split, one per dataset, and it is shared by every method, model, and seed for that dataset. It is the initialization of every alignment run, the denominator of every KL estimate (Appendix~\ref{app:kl}), and the source of the per-response thresholds used by the auxiliary satisfaction metric.

For the two-objective datasets SafeRLHF and AnthropicHH, we train each model with 9 preference vectors $\lambda \in  \left\{\left(\frac{m}{10}, \frac{10-m}{10}\right): m \in [9] \right\}$. For the three-objective SafeRLHF dataset, we train each model on the 15 interior points of the Das-Dennis lattice \citep{dasdennis1998lattice} with $H = 7$, i.e.\ $\lambda \in \left\{ \left( \frac{m}{7}, \frac{n}{7}, \frac{7 - m - n}{7} \right): m \in [5], n \in [5], m + n \le 6 \right\}$. We also train three replicates of each model with different random seeds, for a total of 594 models (162 each for SafeRLHF and AnthropicHH, 270 for HelpSteer2).

\subsection{Sensitivity to Scalarization Hyperparameters $\tau$ and $\gamma$}
\label{app:gamma-tau}
To determine the sensitivity of \method{} to its scalarization hyperparameters $\tau$ and $\gamma$, we replicate the offline off-policy evaluations of Section~\ref{sec:protein-evals}. We train ProGen3-3B using \method{} while sweeping $\tau$ (Table~\ref{tab:tau-sweep}) and $\gamma$ (Table~\ref{tab:gamma-sweep}) using all 3 protein datasets, and we report the hypervolumes of the expected rewards below. We find that performance is stable across a large range of values for both $\tau$ and $\gamma$. However, we do observe some performance degradations on the \textalpha-amylase dataset when $\gamma \ge 0.5$.

\begin{table}[htpb]
    \centering
    \begin{tabular}{|l|ccccc|}
    \hline
    Dataset & $\tau = 0.2$ & $\tau = 0.5$ & $\tau = 1.0$ & $\tau = 2.0$ & $\tau = 5.0$ \\
    \hline
    DHFR & 18.18 & 18.31 & 18.16 & 18.27 & 18.14 \\
    PbrR & 17.73 & 17.96 & 18.21 & 17.26 & 17.42 \\
    \textalpha-Amylase & 27.17 & 27.86 & 30.30 & 26.91 & 26.52 \\
    \hline
    \end{tabular}
    \caption{Hypervolumes sweeping $\tau$ while fixing $\gamma = 0.2$.}
    \label{tab:tau-sweep}
\end{table}

\begin{table}[htpb]
    \centering
    \begin{tabular}{|l|ccccc|}
    \hline
    Dataset & $\gamma = 0.05$ & $\gamma = 0.1$ & $\gamma = 0.2$ & $\gamma = 0.5$ & $\gamma = 1.0$ \\
    \hline
    DHFR & 18.18 & 18.17 & 18.16 & 18.20 & 18.07 \\
    PbrR & 18.04 & 17.72 & 18.21 & 17.97 & 17.35 \\
    \textalpha-Amylase & 26.63 & 27.19 & 30.30 & 22.61 & 20.33 \\
    \hline
    \end{tabular}
    \caption{Hypervolumes sweeping $\gamma$ while fixing $\tau = 1.0$.}
    \label{tab:gamma-sweep}
\end{table}

\section{Generation Hyperparameters}
\subsection{Generated Proteins}
\label{app:plm-generations}
For the DHFR dataset, we select the four DHFR homologs from the training set that maximize the hypervolume. We prompt PG3 with the first 20 residues (out of 161--170 total) on either the N- or C-terminal. For the RA models, we also prepend the prompts used for training. For all models, we autoregressively generate 100,000 proteins via top-$p$ sampling with $p = 0.95$ and temperatures $\tau \in [0.5, 1.0]$ \citep{holtzman2020nucleussampling}, and we apply the quality filtering criteria of \citet{bhatnagar2025progen3}. 

On the other hand, since the PbrR and \textalpha-amylase datasets are local mutational landscapes, any reward model trained on them would be less reliable at evaluating diverse generations from an autoregressive PLM. We instead use the Gibbs with Gradient (GWG) algorithm to propose mutations to the wild type protein using PLM negative log likelihoods as an energy function \citep{grathwohl2021gibbswithgradient}. Because GWG is a Metropolis-Hastings algorithm, we can use the sequences sampled from it to estimate the expected rewards of sampling from the PLM of interest while remaining in-distribution for our reward models \citep{practicalqmc, grathwohl2021gibbswithgradient}. For each model, we run 1500 GWG trajectories for 300 steps each and filter out any sequences with >10 mutations, since these are likely out-of-distribution for our reward models. For the RA models, these trajectories are evenly split across the different training prompts.

\subsection{LLM Generations}
\label{app:llm-generations}
We autoregressively sample one completion per prompt per checkpoint using full ancestral sampling (temperature $\tau = 1.0$, top-$p$ with $p = 1.0$). This was deliberately chosen to properly estimate the KL-divergence from the reference policy to ensure a matched-drift comparison. We sample completions for only 150 prompts for the test sets of all three datasets to account for differing dataset sizes, since AnthropicHH is notably larger than SafeRLHF or HelpSteer2.

\subsection{KL Divergence Matching for Chatbot Alignment Experiments}
\label{app:kl}
\begin{table}[t]
\centering
\small
\begin{tabular}{|ll|rrrr|}
\hline
Dataset & Model & $\eta$ (nats) & Worst dev. & Common range & Width \\
\hline
SafeRLHF  & Llama-3.1-8B & $26.298$ & $0.177$ & $[4.71,\ 27.84]$ & $23.13$ \\
SafeRLHF  & Qwen2.5-7B   & $20.199$ & $1.138$ & $[2.66,\ 25.89]$ & $23.23$ \\
\hline
HH          & Llama-3.1-8B & $16.042$ & $0.405$ & $[2.96,\ 19.42]$ & $16.45$ \\
HH          & Qwen2.5-7B   & $12.890$ & $1.977$ & $[1.89,\ 13.55]$ & $11.66$ \\
\hline
HelpSteer2 & Llama-3.1-8B & 20.616 & 0.471 & [2.12, 21.43] & 19.31 \\
HelpSteer2 & Qwen2.5-7B   & 21.645 & 0.235 & [1.89, 21.53] & 19.64 \\
\hline
\end{tabular}
\caption{Matched-KL budgets.}
\label{tab:kl-budgets}
\end{table}

As discussed in Section~\ref{sec:llm_benchmarks}, we generate one response $y_x \sim \pi(\cdot | x)$ per test prompt for every training checkpoint. We then estimate $\hat{\mathbb{D}}_{\mathrm{KL}}(\pi \mid\mid \pi_0) = \frac{1}{|\mathcal{P}|} \sum_{x \in \mathcal{P}} \log \pi(y_x| x) - \log \pi_0(y_x| x)$. When determining the KL budget $\eta$, we specifically omit checkpoints where the trained policy has drifted less than $1.0$ nats away from the reference policy. We also omit from consideration any $\eta$ that is lower than the one-half the median of the maximum KL achieved by each run to set an informative floor. When there is no $\eta$ above the floor, we instead take $\eta$ as the midpoint of that floor and the lowest maximum-achieved KL in any run. Table~\ref{tab:kl-budgets} describes $\eta$ for each dataset and model, along with what the worst deviation from that budget is for any of the selected checkpoints. The common range of available KL values that can be matched is also described, along with the width of that range.

\section{Multi-Seed Hypervolume Estimation}
\label{app:hv-estimator}
A Pareto front requires one trained policy per preference weight. With $|\Lambda|$ weights and $S$ seeds there are $S^{|\Lambda|}$ ways to assemble one, and seed labels carry no meaning \emph{across} weights, so pairing by index is one arbitrary choice among many and injects variance not present in the data. We therefore report
\begin{equation}
  \widehat{\mathrm{HV}}
  \;=\;
  \frac{1}{S^{|\Lambda|}}
  \sum_{c \in \{1,\dots,S\}^{|\Lambda|}}
  \mathrm{HV}\!\left(\left\{\, r_{\lambda_j,\, c_j} \,\right\}_{j=1}^{|\Lambda|}\right),
  \label{eq:vstat}
\end{equation}
the mean over all assignments $c$ of seeds to weights, where $r_{\lambda,s}$ is the mean reward vector of the policy trained at weight $\lambda$ with seed $s$.

Standard errors come from a bootstrap over the unit of replication: for each preference weight, we independently resample its $S$ seeds with replacement, recompute Equation~\ref{eq:vstat}, and repeat $S^{|\Lambda|}$ times. We take this approach because enumerating all $S^{|\Lambda|}$ assignments artificially decreases the reported standard error for the LLM experiments, which have higher variance because they rely on just one completion per prompt per checkpoint.

Two methods are compared by bootstrapping each independently and reporting the one-sided proportion of replicates in which the difference does not favor the nominal leader. Tables mark every method not separated from the leader at the $5$\% level. Such a mark says a method has \emph{not been shown to be worse}.

In all settings, we compute hypervolumes in z-score units. We set the reference point as the minimum observed reward independently over rewards across the train and test splits.

\section{Gaussian Process Reward Models}
\label{app:gp-reward-models}
We wish to gauge the generative capabilities of the protein language models when trained with different optimization objectives across different base models and datasets as described in Section~\ref{sec:protein-evals}. 
For this, we turn to the Bayesian optimization literature, where probabilistic surrogate models are commonly used to estimate the utility of observing new data points~\citep{freitas2016bayesopt}.

By conditioning a Gaussian Process (GP) \citep{rasmussen2005gp} surrogate model on the experimental data, we obtain a posterior predictive distribution over the sampled sets of generated sequences.
We can then compute the expected hypervolumes and use these as indicators of the relative generative capabilities of the aligned models.

We model the experimental outcomes as $y=f(\mathbf{x})+\epsilon$, where $f$ is some unknown function over protein sequences and $\epsilon\sim\mathcal{N}(0,\sigma_\epsilon^2)$ is Gaussian distributed noise.
We place a GP prior over the function $f$, such that
\begin{align*}
    f(\mathbf{x})\sim\mathcal{G{P}}(m,k(\mathbf{x}, \mathbf{x'})).
\end{align*}
Our GP prior is characterized by a constant mean $m$ and a Mat{\'e}rn 5/2 kernel \citep{rasmussen2005gp}:
\begin{align*}
    k_{\text{Mat{\'e}rn}\frac{5}{2}}(\mathbf{x}, \mathbf{x}')=\left(1+\frac{\sqrt{5}\|\mathbf{x}-\mathbf{x}'\|_2}{\ell}+\frac{5\|\mathbf{x}-\mathbf{x}'\|_2^2}{3\ell^2}\right)\exp\left(-\frac{\sqrt{5}\|\mathbf{x}-\mathbf{x}'\|_2}{\ell}\right)
\end{align*}
We embed all protein sequences using the E1-600m model~\citep{jain2025e1} and apply mean pooling across the length dimension to obtain 1280 dimensional embeddings. 
We follow \citet{hvarfner2024vanilla} and place a LogNormal ($\mathcal{LN}$) hyperprior over the lengthscale of the Mat{\'e}rn kernel:
\begin{align*}
    \ell \sim \mathcal{LN}\left(\mu_0+\frac{\log(D)}{2},\sigma_0\right),
\end{align*}
with $\mu_0=\sqrt{2}$, $\sigma_0=\sqrt{3}$, and $D=1280$, without applying automatic relevance determination over embedding dimensions which degraded performance.
We standardize all target variables independently to both improve hyperparameter optimization stability and to prevent particular targets from dominating the hypervolumes. 
We do not fix the signal variance $\sigma_f^2$ to $1.0$ as in \citet{hvarfner2024vanilla} which led to worse performance on a held-out test set but treat it as a learnable kernel hyperparameter which is multiplied with the Mat{\'e}rn kernel such that our final kernel formulation is 
\begin{align*}
k(\mathbf{x}, \mathbf{x}')=\sigma_f^2\,k_{\text{Mat{\'e}rn}\frac{5}{2}}(\mathbf{x}, \mathbf{x}').
\end{align*}
We place a Gamma prior with a mean of $1.0$ on the signal variance, $\sigma_f^2 \sim \Gamma\left(\alpha, \beta\right)$, with shape parameter $\alpha=5$ and rate parameter $\beta=5$, providing adequate regularization.
For the constant mean $m$, we use a Gaussian prior such that $m\sim\mathcal{N}(0,\sigma_m^2)$ with $\sigma_m=3$.
Lastly, we use a default BoTorch prior for the noise variance such that $\sigma_\epsilon^2\sim\Gamma(1.1, 0.5)$ \citep{botorch}.

For each dataset, we fit independent GPs to each target by optimizing the log marginal likelihood with an L-BFGS-S optimizer in double precision using BoTorch~\citep{botorch}.
We fit the surrogate models to the full experimental datasets (whereas the PLMs were only trained on a subset of the data) to maximize the support and predictive power of the GPs as oracles.

Instead of averaging the activity across codons as a preprocessing step for the DFHR dataset, we train GPs for each activity/codon combination for a total of four models.
For each prediction, we then take the average of each codon-specific GP.
If a given sequence only has an experimental measurement for a single codon in one of the two activity objectives (as an example), we can still use it instead of requiring experimental measurements for all four activity/codon combinations.
This way, we maximize the number of training sequences the oracle can leverage.

\subsection{Reward model validation}
\label{app:gp-eval}
\begin{table}[tbp]
    \centering
    \begin{tabular}{|ll|ccc|}
        \hline
        Dataset & Target & Spearman & NCDG@10\% & MSE \\
        \hline
        PbrR & Pb\_log2FC\_increase & 0.789 & 0.979 & 0.422 \\
        PbrR & Zn\_log2FC\_decrease & 0.800 & 0.858 & 0.277 \\
        \hline
        DHFR & 0{\textmu}g/mL\_TMP\_log2FC & 0.503 & 0.947 & 0.587 \\
        DHFR & 50{\textmu}g/mL\_TMP\_log2FC & 0.573 & 0.813 & 0.960  \\
        \hline
        \textalpha-Amylase & activity\_log2FC & 0.581 & 0.895 & 1.296  \\
        \textalpha-Amylase & stability\_log2FC & 0.517 & 0.874 & 1.823 \\
        \textalpha-Amylase & expression\_log2FC & 0.510 & 0.868 & 0.479 \\
        \hline
    \end{tabular}
    \caption{Predictive performance of GP regression models on held-out test data.}
    \label{tab:surrogate-evals}
\end{table}

To validate the use of the Gaussian process regression models for generative evaluations, we fit them to a subset of each dataset and evaluate their predictive performance on held-out test sets. 
As a result of the iterative creation of the PbrR and \textalpha-Amylase datasets, the mutation order varies substantially from the original training to test sets.
The training partitions primarily consist of single mutations to a wild type, whereas the subsequently collected test sets contain upwards of 10 mutations.
For the generative evaluations, we use all experimental data to train the oracle models while limiting the number of mutations in the generated sequences to 10 as described in Section~\ref{sec:protein-evals}.
To approximate the training and inference distributions encountered in the generative evaluations as closely as possible, we therefore randomly place half of the designated test sequences in the training set.

We report the Spearman rank correlation, NCDG at 10\% as well as the mean squared error across the three datasets and seven targets in Table~\ref{tab:surrogate-evals}.
We observe good predictive performance for all configurations, justifying the use of the Gaussian process regressors as reward models for the generative evaluations.

\subsection{Hypervolume estimation}
\label{app:hypervolume}
To compute the expected hypervolume for a given set of $k$ sequences, we rely on quasi-Monte Carlo (QMC) integration by drawing 256 samples from the posterior predictive distribution using Sobol sequences.
For all QMC samples, we compute the resulting exact hypervolumes using the algorithms described in \citet{hvalgo1,hvalgo2,hvalgo3} via the \texttt{moocore} library \citep{moocore} and take the average.
We repeat this process for each $k \in \{12, 24, 48, 96, 192, 384\}$ a total of 100 times with different sampled sets to reduce variance and estimate uncertainty.
For each dataset, we select a fixed reference point for hypervolume computations as the minimum observed reward independently across rewards.

\section{Re-parameterization of Energy Rank Alignment}
\label{app:era}
In its original formulation, ERA introduces two parameters, $\beta_{\mathrm{ERA}}$ for sample diversity and $\gamma_{\mathrm{ERA}}$ for reference policy regularization, that carry different semantic meanings compared to their usual counterparts in the RL and preference optimization literature~\cite{chennakesavalu2025aligning}. In order to translate to the conventional definition of $\beta$ referenced throughout this work, we first take the definition of the optimal target distribution as follows:
\begin{align}
\pi^\star(y\mid x) &= Z^{-1}(x)\exp\!\left(-\tfrac{\beta_{\mathrm{ERA}}}{1+\gamma_{\mathrm{ERA}}}\big(U(x,y) - \beta_{\mathrm{ERA}}^{-1}\gamma_{\mathrm{ERA}}\log\pi_{0}(y\mid x)\big)\right) \label{eq:era-min}
\end{align}
where $\pi_{0}$ is the reference policy and $U$ is the energy to be minimized (negative reward in this setting). Consistent with the original work, we define the following bijection to re-parameterize ERA: 
\begin{equation}
\beta \;=\; \frac{1+\gamma_{\mathrm{ERA}}}{\beta_{\mathrm{ERA}}}, \qquad
w \;=\; \frac{\gamma_{\mathrm{ERA}}}{1+\gamma_{\mathrm{ERA}}} \in [0,1)
\label{eq:reparam}
\end{equation}
We introduce the $w$ term as a measure of \textit{the extent of regularization towards the reference policy versus a uniform distribution}, i.e.\ maximum-entropy regularization. When framed in the context of the RL Optimization Problem~\ref{eq:morl}, the ERA objective can be defined as follows:
\begin{equation}
\max_{\pi}\; \E_{x\sim\mathcal{P}}\,\E_{y\sim\pi(\cdot\mid x)}\big[r(x,y)\big]
\;-\; \beta\Big[\, w\, \mathbb{D}_\mathrm{KL}(\pi \mid\mid \pi_{0}\big) \;+\; (1-w)\, \mathbb{H}(\pi)\Big]
\label{eq:era_lagrangian}
\end{equation}
The reparameterized Equation~\ref{eq:era_lagrangian} now adopts the field-standard definition of $\beta$ for fair comparison with other baseline algorithms, and makes clear the role of $w$ under this parameterization.

\iftoggle{anonymous}{
%\clearpage
%\input{checklist}
}{}

\end{document}